\documentclass[11pt]{article}
\usepackage{float}
\usepackage{graphicx}
\usepackage{amsmath, amssymb}
\usepackage{booktabs}
\usepackage{authblk}
\usepackage{comment}
\usepackage[margin=1in]{geometry}

\title{\bfseries WIPUNet: A Physics-inspired Network with Weighted Inductive Biases for Image Denoising}

\author[1]{\textbf{Wasikul Islam}}
\affil[1]{Department of Physics, University of Wisconsin, Madison, WI 53706 USA}

\date{}

\begin{document}
\maketitle

\begin{abstract}
In high-energy particle physics, collider measurements are contaminated by 
``pileup''---overlapping soft interactions that obscure the hard-scatter 
signal of interest. Dedicated subtraction strategies exploit physical 
priors such as conservation, locality, and isolation. Inspired by this 
analogy, we investigate how such principles can inform image denoising by 
embedding physics-guided inductive biases into neural architectures. 
\textbf{This paper is a proof of concept}: rather than targeting state-of-the-art (SOTA) benchmarks, 
we ask whether physics-inspired priors improve \emph{robustness} under strong corruption.

We introduce a hierarchy of PU-inspired denoisers: a residual CNN with conservation constraints, 
its Gaussian-noise variants, and the \textbf{Weighted Inductive Pileup-physics-inspired U-Network for 
Denoising (WIPUNet)}, which integrates these ideas into a UNet backbone. 
On CIFAR-10 with Gaussian noise at $\sigma\in\{15,25,50,75,100\}$, PU-inspired CNNs are competitive with 
standard baselines, while WIPUNet shows a \emph{widening margin} at higher noise. 
Complementary BSD500 experiments show the same trend, suggesting physics-inspired priors provide 
stability where purely data-driven models degrade. Our contributions are: (i) translating pileup-mitigation 
principles into modular inductive biases; (ii) integrating them into UNet; and (iii) demonstrating robustness 
gains at high noise without relying on heavy SOTA machinery.
\end{abstract}

\section{Introduction}
Image denoising remains a central benchmark in computer vision, with standard baselines including 
DnCNN~\cite{zhang2017beyond}, FFDNet~\cite{zhang2018ffdnet}, UNet~\cite{ronneberger2015unet}, and 
Restormer~\cite{zamir2022restormer}. These architectures learn purely from image statistics, typically 
optimized for additive white Gaussian noise (AWGN). While effective at mild corruption, many CNN/Transformer 
denoisers \emph{degrade as noise increases}, and they provide limited physical interpretability about 
why certain structures are preserved or suppressed.

In contrast, high-energy particle physics (HEP) faces an analogous but physically grounded problem: 
disentangling the ``hard-scatter'' signal from diffuse pileup (PU)---tens of overlapping proton--proton 
collisions occurring in each bunch crossing. Experimental collaborations at CERN such as ATLAS and CMS have 
developed mitigation strategies grounded in physically motivated priors:
\begin{itemize}
    \item \textbf{Jet area–based subtraction:} event-by-event estimation of pileup transverse momentum density 
    $\rho$, subtracted according to jet area to correct jet energy and shapes~\cite{ATLAS2015_PileupMitigation}.
    \item \textbf{Constituent-level and grooming techniques:} extensions such as Voronoi subtraction and tools 
    like SoftKiller, Constituent Subtraction, or PUPPI enable particle-level pileup removal with improvements 
    at high collision rates~\cite{Berta2019_ICS, CMS2020_PileupMitigation,PUPPI2014}.
    \item \textbf{Machine learning–based classification:} PUPPI and graph neural networks suppress pileup by 
    estimating the origin of each particle~\cite{PUPPI2014,Cerri2018_GNN}.
\end{itemize}

This work draws a cross-disciplinary bridge: reinterpreting pileup subtraction as image denoising. 
By embedding physics priors into neural architectures, we develop PU-inspired denoisers that are competitive 
with established baselines while offering \emph{interpretability} and improved \emph{robustness} under heavy noise.

In this work, we introduce a set of physics-inspired inductive biases and map them to modular network components that can be plugged into a deep learning backbone. We integrate these modules into a UNet architecture, which allows both ablation of individual principles and their combined effect. Finally, we demonstrate on CIFAR-10~\cite{krizhevsky2009cifar10} and BSD500~\cite{MartinFTM01} that these physics-guided priors provide increasing robustness as noise intensifies, underscoring the role of this study as a \emph{proof of concept} rather than a state-of-the-art benchmark.

\section{Why CIFAR-10 is a Meaningful Testbed for Pileup-Inspired Denoising}
The CIFAR-10 dataset~\cite{krizhevsky2009cifar10}, though often viewed as a ``toy'' benchmark, offers a 
\emph{low-redundancy} regime that is \emph{stringent} for denoising. Each $32\times32$ RGB image contains few 
pixels per object; small corruptions can erase entire semantic parts (e.g., a bird beak). Compared to larger datasets 
(e.g., ImageNet $224\times224$), CIFAR-10 provides less spatial repetition to average out noise, making robustness 
and priors more consequential.

Under AWGN with $\sigma=15$, single-pixel perturbations can disproportionately affect structure. In such low-redundancy 
settings, \emph{conservation} (residual subtraction), \emph{$\sigma$-conditioning}, \emph{isolation} (attention), and 
\emph{multi-scale} reasoning directly influence whether informative features are retained. This mirrors HEP scenarios where sparse measurements demand strong constraints. 

We therefore \emph{deliberately} use CIFAR-10 to stress-test priors and then validate the observed trends on the higher-resolution, natural-image BSD500 dataset, which serves as a complementary check beyond tiny images.

\begin{figure}[ht]
\centering
\begin{tabular}{cccc}
\includegraphics[width=0.3\linewidth]{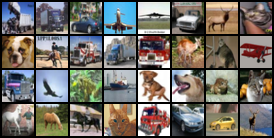} &
\includegraphics[width=0.3\linewidth]{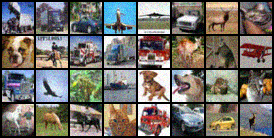} &
\includegraphics[width=0.3\linewidth]{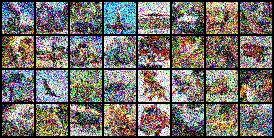} \\

(a) Clean & (b) Noisy ($\sigma$=15) & (c) Noisy ($\sigma$=100) \\[0.5em]

\end{tabular}
\caption{Examples of a few CIFAR-10 images, demonstrating Gaussian corruption at low ( $\sigma=15$) and very high noise levels ($\sigma=100$).}
\label{fig:cifar10_noisy_images}
\end{figure}

\section{Particle-Physics Inspired Design Principles}

In high-energy collider experiments, pileup mitigation has motivated a rich set 
of inductive biases that are equally relevant for designing robust learning 
architectures. Detector measurements can be viewed as mixtures of signal and 
background contributions, requiring reconstruction algorithms that (i) conserve 
global quantities, (ii) distinguish structured, localized features from diffuse 
backgrounds, (iii) isolate salient objects from surrounding contamination, and 
(iv) adapt performance based on external conditions such as pileup multiplicity. 
These guiding principles have direct analogies in image denoising, where the 
goal is to recover structured content from noisy observations. 

We translate them into four modular design strategies for denoising: 
\emph{(i) conservation via residual subtraction}, 
\emph{(ii) noise-level conditioning via a $\sigma$-map}, 
\emph{(iii) isolation via channel attention}, and 
\emph{(iv) multi-scale fusion via learned resampling}. 
Each module is implemented as a lightweight, pluggable component within a UNet 
backbone, enabling ablation of their isolated and combined effects. 
We now describe each principle in detail.

\subsection{Hard conservation via residual learning.}
Rather than directly predicting the clean image, the network learns the noise 
field $\hat{N}$ and subtracts it from the input:
\[
\hat{I}_{\mathrm{clean}} = I_{\mathrm{noisy}} - \hat{N}.
\]
This formulation enforces that output intensities remain consistent with the 
input, mirroring how energy–momentum conservation constrains subtraction 
techniques in particle physics.  
\textbf{Example in HEP:} In vertex reconstruction, multiple proton--proton 
collisions (pileup) produce extra tracks in addition to those from the 
hard-scatter interaction. Algorithms assign tracks to either the 
hard-scatter or pileup vertices, and when pileup tracks are removed, 
momentum conservation must still hold: the vector sum of transverse momenta 
($\sum \vec{p}_T$) at the hard-scatter vertex should remain consistent with 
the underlying particle physics interaction.

\subsection{Noise-level conditioning with a $\sigma$-map.}
The model receives an auxiliary channel encoding the noise variance 
$\sigma(x,y)$, enabling explicit conditioning on the corruption level:
\[
f_\theta \big(I_{\mathrm{noisy}}, \sigma(x,y)\big) \;\to\; \hat{I}_{\mathrm{clean}}.
\]
This is analogous to conditioning physics reconstruction on the measured pileup 
multiplicity $N_{\mathrm{PU}}$, which strongly influences background 
fluctuations.  
\textbf{Example in HEP:} Jet energy corrections incorporate the pileup density 
$\rho$ as an external observable to improve background subtraction.

\subsection{Isolation via Squeeze-and-Excitation attention.}
To emphasize informative feature channels while suppressing spurious ones, we 
employ Squeeze-and-Excitation (SE) blocks that apply global importance weights:
\[
\tilde{F}_c = s_c \cdot F_c, \quad 
s_c = \sigma\!\left( W_2 \, \delta\!\left( W_1 \, \mathrm{GAP}(F) \right)\right),
\]
where $\mathrm{GAP}$ denotes global average pooling. This channel-wise reweighting 
acts as an isolation criterion.  
\textbf{Example in HEP:} Isolated high-$p_T$ physics objects are selected as signal-like, 
while diffuse soft activity from pileup is suppressed.

\subsection{Multi-scale learned resampling.}
Instead of fixed pooling and interpolation, we implement downsampling and 
upsampling through residual blocks with learnable filters. This enables the 
network to capture correlations across multiple spatial scales.  
\textbf{Example in HEP:} Detector subsystems operate at different resolutions: 
trackers provide fine spatial detail while calorimeters integrate energy over 
larger regions. Combining information across scales is crucial for robust 
object reconstruction under high pileup conditions.

\section{PU-Inspired Models}
We introduce three progressively more sophisticated physics-inspired denoisers.

\subsection{Simple-PU-CNN}
Simple-PU-CNN is a residual CNN inspired by DnCNN but modified to predict background $\hat{B}$ rather than the clean signal directly. Conservation is enforced by subtraction:
\begin{equation}
    \hat{S} = Y - \hat{B}.
\end{equation}
This design mirrors pileup subtraction strategies, ensuring that denoising respects mixture conservation.

\subsection{PU-Net-G and PU-Net++}
Two Gaussian variants extend Simple-PU-CNN:

\paragraph{PU-Net++.} 
PU-Net++ is designed as an explicit \emph{mixture model}, inspired by how collider physics separates hard-scatter signal from diffuse pileup. A compact U-Net backbone produces three auxiliary maps: a non-negative background density $\rho$ (via Softplus), a binary-like signal mask $m$ (via Sigmoid), and a confidence gate $g$ (via Sigmoid). The clean estimate is formed as $S = (g \cdot m)\,(Y-\rho)$, with the background $B$ obtained by hard conservation $B=Y-S$. This decomposition reflects key physics priors: (i) \textbf{mixture conservation}, enforcing $Y=S+B$; (ii) a \textbf{density prior}, modeling pileup as a diffuse positive field; and (iii) \textbf{isolation}, where $m$ and $g$ suppress pileup-like features while retaining localized, structured signal regions.

\paragraph{PU-Net-G.} 
PU-Net-G takes a complementary approach, aligning with residual-based subtraction methods used in pileup mitigation. Instead of modeling density or masks, the network directly predicts the noise field $\hat{N}$, subtracting it from the input: $S=Y-\hat{N}$. A one-channel $\sigma$-map is concatenated to the RGB input, conditioning the model on the noise level---analogous to conditioning on pileup multiplicity in collider physics. Its design reflects three physics-inspired priors: (i) \textbf{hard conservation}, guaranteeing consistency by subtraction; (ii) \textbf{noise-level conditioning}, guiding the denoiser with an external observable (the corruption strength $\sigma$); and (iii) \textbf{multi-scale resampling}, using learned encoder--decoder pathways to capture both fine and coarse correlations, much like combining tracker and calorimeter information. 

Both are competitive with strong CNN baselines on Gaussian noise, validating the physics-inspired formulation.

\subsection{WIPUNet: Embedding Physics Priors into UNet}

To move beyond baseline denoising, we embed particle-physics-inspired priors 
into a UNet backbone, producing the family of models we collectively call \textbf{Weighted Inductive Pileup-physics-inspired U-Network for Denoising} or 
\textbf{WIPUNet}. Each variant incorporates a different inductive bias 
motivated by pileup mitigation in collider physics.

\paragraph{Physics Motivation.}
In high-luminosity colliders, pileup background $B$ is diffuse in 
$\eta$--$\phi$, while signal $S$ is structured and subject to 
conservation laws. WIPUNet encodes these differences explicitly through:
\begin{itemize}
    \item \emph{Residual subtraction:} enforces $S = Y - \hat{B}$, mirroring 
    the mixture model $Y=S+B$.
    \item \emph{Conditioning:} auxiliary $\sigma$-maps provide noise-level or 
    multiplicity information.
    \item \emph{Isolation:} attention mechanisms suppress diffuse features, 
    analogous to lepton/jet isolation.
    \item \emph{Multi-scale fusion:} learned resampling captures correlations 
    across scales, akin to combining tracker and calorimeter data.
\end{itemize}

\paragraph{Variants.}
To disentangle the contribution of each principle, we define multiple WIPUNet
variants:
\begin{itemize}
    \item \textbf{WIPUNet1 (Hard conservation).}  
    Learns to predict noise and subtract it.  
    \emph{Physics analogy:} momentum-conserving pileup subtraction.
    \item \textbf{WIPUNet2 (Noise-level conditioning).}  
    Input includes a $\sigma$-map channel.  
    \emph{Physics analogy:} conditioning on pileup multiplicity.
    \item \textbf{WIPUNet3 (SE attention).}  
    Adds channel-wise attention to highlight informative features.  
    \emph{Physics analogy:} isolation of signal-like objects.
    \item \textbf{WIPUNet4 (Multi-scale resampling).}  
    Replaces pooling with ResBlocks in down/up paths.  
    \emph{Physics analogy:} combining fine-grained tracker with coarse calorimeter.
    \item \textbf{WIPUNet.}  
    Integrates all four physics-inspired modules.  
    \emph{Physics analogy:} full pileup mitigation stack in collider analyses.
\end{itemize}

\paragraph{Architecture.}
All models retain the encoder–decoder UNet backbone. Physics-inspired modules 
are inserted modularly: the residual head enforces conservation, the auxiliary 
channel encodes $\sigma$, SE blocks reweight skip features, and ResBlocks 
replace pooling for multi-scale correlations.

\paragraph{Training Objective.}
The loss combines direct reconstruction and residual background consistency:
\begin{equation}
    \mathcal{L} = \lambda_{\text{img}} \cdot \| \hat{S} - S \|_2^2 \;+\;
    \lambda_{\text{res}} \cdot \| \hat{B} - (Y-S) \|_2^2,
\end{equation}
ensuring both signal fidelity and physically consistent background estimation.


\begin{figure}[H]
    \centering  \includegraphics[height=1.2\textwidth]{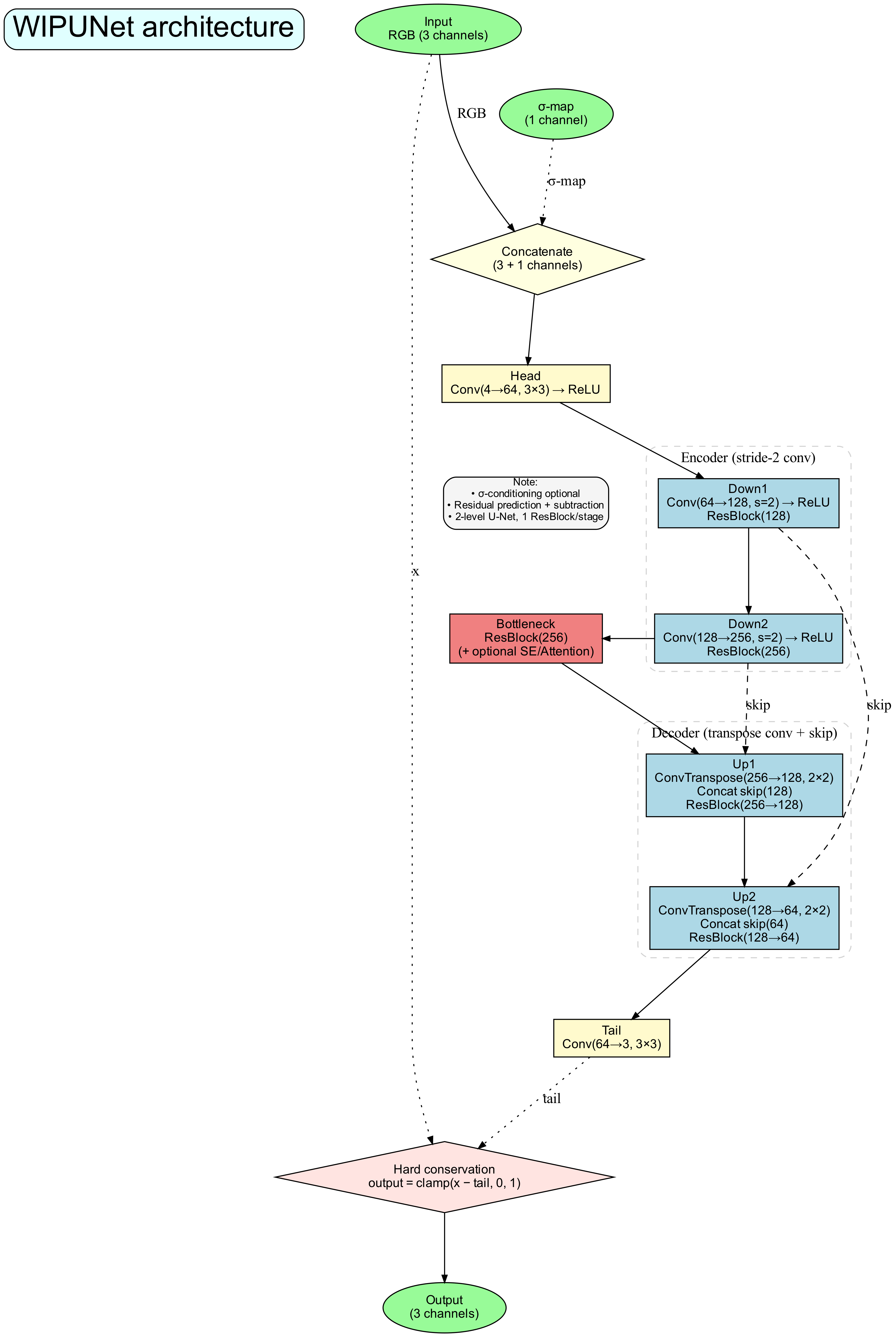}
    \caption{
    Conceptual view of \textbf{WIPUNet}: a UNet backbone augmented with four physics-inspired modules. 
    (1) \emph{Residual head} enforces conservation ($\hat S=Y-\hat N$); 
    (2) a \emph{$\sigma$-map} channel conditions on noise level; 
    (3) \emph{SE attention} isolates informative channels on skips; and 
    (4) \emph{learned resampling} replaces fixed pooling/upsampling to fuse multi-scale correlations..}
    \label{fig:WIPUNet_archi}
\end{figure}

\section{Training and Evaluation Details}
\label{sec:setup}

\paragraph{Dataset and splits.}
All experiments use CIFAR-10 (50k train, 10k test). Images are RGB,
$32\times 32$. We do not apply geometric or color augmentations; the
only stochasticity comes from noise injection.

\paragraph{Noise model and protocol.}
For a target level $\sigma\in\{15,25,50, 75, 100\}$, additive white Gaussian noise
(AWGN) is injected \emph{on-the-fly} during training and evaluation:
$Y = S + \varepsilon$, with $\varepsilon\sim\mathcal{N}(0, (\sigma/255)^2)$.
For $\sigma$-aware models (FFDNet, PU variants with conditioning), we
concatenate an additional single-channel \emph{$\sigma$-map} of constant
value $\sigma/255$ to the RGB input (shape $[B,1,H,W]$).

\paragraph{Optimization.}  
We train for \textbf{100 epochs} using \textbf{AdamW} (learning rate 
$5\times 10^{-4}$, default $\beta$'s, AdamW default \emph{weight decay}). 
Mixed-precision (AMP) is enabled on CUDA. Gradients are \textbf{clipped to 1.0}. 
The random seed is fixed to $1234$ for reproducibility.

\paragraph{Batching and data loading.}  
Default batch size is \textbf{adaptive} (512 on CUDA/MPS, 64 otherwise), with 
\texttt{num\_workers} $\geq 2$ selected via an auto-heuristic. For heavier 
models the batch size is reduced if VRAM is limited.

\paragraph{Losses.}  
All models---including DnCNN, FFDNet, UNet, RestormerLite, PU-Net-G, and 
WIPUNet (PU\_UNet\_1234)---are trained with an $\ell_2$ image reconstruction loss:  
\[
\mathcal{L} = \| \hat S - S \|_2^2.
\]  
Although PU-Net-G and WIPUNet implement \emph{residual subtraction} internally 
(predicting the noise field $\hat N$ and returning 
$\hat S = Y - \hat N$), the training loss is applied only to the denoised 
output $\hat S$; no separate noise-prediction loss is used. PU-Net++ is trained 
identically, without additional regularizers in this implementation.

\paragraph{Evaluation.}  
We report mean \textbf{PSNR (dB)} and \textbf{SSIM} on the CIFAR-10 test set 
(10k images), computed in PyTorch without external dependencies. For 
$\sigma$-aware models, the matching $\sigma$-map is provided at test time. 
We focus on PSNR/SSIM to isolate the effect of the proposed priors; 
perceptual metrics are left to future work. Exponential Moving Average (EMA) is not used.

\paragraph{Implementation details.}  
$\sigma$-aware models (FFDNet, PU-Net-G, WIPUNet) receive a \textbf{1-channel 
$\sigma$-map} concatenated to the RGB input. DnCNN, UNet, and RestormerLite 
operate only on RGB. All models perform residual subtraction inside the 
forward pass, i.e.\ they predict the noise and return $Y - \hat N$.

\section{Results}

Our emphasis is on \emph{robustness under increasing noise} rather than advancing SOTA benchmarks; we therefore compare against widely used CNN baselines and UNet to test whether the proposed priors stabilize performance as $\sigma$ grows.

We evaluate on CIFAR-10 corrupted with Gaussian noise at multiple levels 
($\sigma=15,25,50,75, 100$). Table~\ref{tab:allresults} reports PSNR and SSIM 
for standard denoisers, UNet, and PU-inspired variants.

\begin{table}[ht]
\centering
\begin{tabular}{lcccccccccc}
\toprule
 & \multicolumn{2}{c}{$\sigma=15$} & \multicolumn{2}{c}{$\sigma=25$} & 
 \multicolumn{2}{c}{$\sigma=50$} & \multicolumn{2}{c}{$\sigma=75$} & \multicolumn{2}{c}{$\sigma=100$} \\
Model & PSNR & SSIM & PSNR & SSIM & PSNR & SSIM & PSNR & SSIM & PSNR & SSIM\\
\midrule
DnCNN & 26.51 & 0.892 & 28.67 & 0.936 & 24.90 & 0.866 & 22.76 & 0.800 & 21.49 & 0.765 \\
FFDNet & 31.91 & 0.969 & 28.94 & 0.941 & 25.00 & 0.871 & 22.87 & 0.811 & 21.23 & 0.757\\
RestormerLite & 28.48 & 0.934 & 24.91 & 0.867 & 20.33 & 0.724 & 18.01 & 0.617 & 16.67 & 0.544\\
UNet & 31.95 & 0.969 & 29.03 & 0.941 & 25.28 & 0.878 & 23.11 & 0.819 & 21.62 & 0.773\\
\midrule
PUNet++ & 24.80 & 0.853 & 20.49 & 0.721 & 14.94 & 0.496 & 12.06 & 0.366 & 10.33 & 0.289\\
PUNetG & 31.50 & 0.965 & 29.30 & 0.946 & 25.68 & 0.889 &23.65 & 0.836 & 22.21 & 0.788\\
WIPUNet & \textbf{32.05} & \textbf{0.969} & \textbf{29.32} & \textbf{0.946} & \textbf{25.75} & \textbf{0.890} & \textbf{23.71} & \textbf{0.837} & \textbf{22.30} & \textbf{0.792}\\
\bottomrule
\end{tabular}
\caption{Denoising performance on CIFAR-10 across Gaussian noise levels $\sigma=15,25,50,75,100$ (left to right). Higher is better. WIPUNet is compared against standard baselines and the highest values are highlighted by bold font.}
\label{tab:allresults}
\end{table}

\begin{figure}[h]
\centering
\begin{tabular}{c}
\includegraphics[width=\linewidth,height=0.3\textheight]{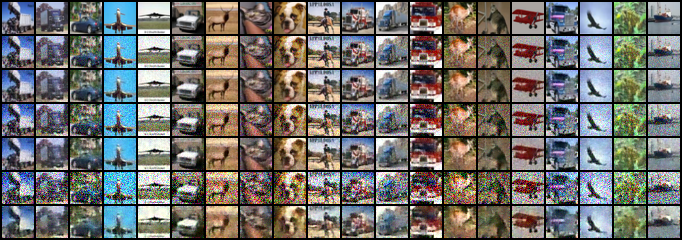} \\
\end{tabular}
\caption{Qualitative denoising results on CIFAR-10 using WIPUNet. 
The \textbf{top row} shows clean reference images. The \textbf{second row} shows 
the same images corrupted with Gaussian noise at $\sigma=15$, followed by 
their WIPUNet denoised outputs in the \textbf{third row}. The \textbf{fourth 
and fifth rows} show noisy and denoised versions for $\sigma=25$, while the 
\textbf{sixth and seventh rows} show noisy and denoised versions for 
$\sigma=50$ respectively. WIPUNet successfully restores fine details and 
object structure even under increasing noise levels.}
\label{fig:cifar10_denoising}
\end{figure}



\begin{table}[h]
\centering
\begin{tabular}{lcccccccccc}
\toprule
 & \multicolumn{2}{c}{$\sigma=15$} & \multicolumn{2}{c}{$\sigma=25$} & \multicolumn{2}{c}{$\sigma=50$} &
 \multicolumn{2}{c}{$\sigma=75$} &
 \multicolumn{2}{c}{$\sigma=100$} \\
Model & PSNR & SSIM & PSNR & SSIM & PSNR & SSIM & PSNR & SSIM & PSNR & SSIM\\
\midrule
DnCNN & 24.85 & 0.551 & 20.12 & 0.377 & 18.08 & 0.337 & 19.01  &  0.300  & 16.44 & 0.240\\
FFDNet & 24.79 & 0.539 & 28.24 & 0.730 & 16.08 & 0.227 & 12.88  &  0.160  & 11.82  & 0.127\\
UNet & 31.47 & 0.827 & 28.56 & 0.737 & 24.85 & 0.607 & 23.00 &  0.512   & 21.86  & 0.450\\
RestormerLite & 29.81 & 0.860 & 27.13 & 0.787 & 23.53 & 0.631 & 22.54  & 0.452  &  21.68  &  0.418 \\
PUNetG & 28.76 & 0.719 & 27.12 & 0.682 & 21.16 & 0.489 & 22.46 &  0.453  &  21.08 & 0.387\\
WIPUNet & 30.94 & 0.814 & 28.01 & 0.735 & \textbf{25.19} & 0.584 & \textbf{23.57}  &  0.507  &  \textbf{23.10} & \textbf{0.467}\\
\bottomrule
\end{tabular}
\caption{Denoising performance results on BSD500 dataset across different noise levels.}
\label{table:sync_bsd500_results}
\end{table}


The consolidated results (Table~\ref{tab:allresults}) highlight several 
important insights into the role of physics-inspired priors for image 
denoising:

\begin{itemize}
\item \textbf{Low noise regime ($\sigma=15$).} WIPUNet slightly surpasses UNet in PSNR while matching its SSIM, indicating that physics-inspired inductive biases can compete with state-of-the-art architectures even under mild corruption. 
    
\item \textbf{Moderate noise regime ($\sigma=25$).} WIPUNet modestly outperforms both UNet and FFDNet (by $\sim$0.3 dB PSNR and 0.005 SSIM), while significantly surpassing RestormerLite. This suggests that conservation and conditioning modules help maintain signal quality under moderate corruption, though the margin over strong CNN baselines is not large.

\item \textbf{High noise regime ($\sigma=50$).} The resilience of WIPUNet becomes more apparent: while all classical baselines degrade substantially, WIPUNet continues to deliver the best results, exceeding UNet by $\sim$0.5 dB PSNR and 0.012 SSIM. This demonstrates that physics priors confer robustness under stronger corruption, where purely data-driven CNNs deteriorate.

\item \textbf{Very high noise regime ($\sigma=75$).} WIPUNet consistently outperforms all baselines, surpassing UNet by $\sim$0.6 dB PSNR and 0.018 SSIM. Even as corruption intensifies, physics-inspired modules stabilize learning and preserve structure better than purely data-driven models.

\item \textbf{Extreme noise regime ($\sigma=100$).} WIPUNet maintains a clear advantage, exceeding UNet by $\sim$0.7 dB PSNR and 0.019 SSIM, while RestormerLite and DnCNN collapse to much lower values. The widening gap at this regime highlights the robustness of hybrid physics–deep learning strategies under severe corruption. 

\item \textbf{Failure of weak baselines.} PUNet++ collapses at higher $\sigma$, confirming that physics priors alone are insufficient without a strong backbone. In contrast, PU-Net-G and WIPUNet maintain strong performance, validating the hybrid strategy of combining domain priors with expressive architectures. 

\end{itemize}

\paragraph{PU-Net++ vs.~PU-Net-G.}
A sharp divergence emerges between these two models. PU-Net++ explicitly implements mixture decomposition ($Y=S+B$), with modules for density estimation, gating, and masking. While conceptually aligned with pileup mitigation in physics, its lightweight UNet backbone proves insufficient for complex natural images, leading to poor performance across all noise levels (PSNR $24.8/20.5/14.9/12.1/10.3$ for $\sigma=15/25/50/75/100$). PU-Net-G, on the other hand, leverages two key physics-inspired mechanisms: residual subtraction (hard conservation) and $\sigma$-conditioning. This combination yields results on par with UNet across all regimes (PSNR $31.5/29.3/25.7/23.7/22.2$), demonstrating that conditioning on noise level and multi-scale resampling provide tangible benefits. The comparison suggests that while explicit mixture modeling offers interpretability, robust performance requires hybrid approaches that embed physics priors into powerful backbones.

To assess the contribution of individual physics-inspired modules, we evaluated WIPUNet variants (WIPUNet1–4), each implementing a single inductive bias, against the full WIPUNet model that integrates all four. Results show that while individual modules provide incremental gains, their combination in WIPUNet yields superior denoising performance, highlighting the complementary nature of these physics principles.

\paragraph{BSD500 experiments.}
To verify that our findings are not CIFAR-specific, we trained all models
from scratch on BSD500~\cite{MartinFTM01} and evaluated on its held-out 
test set across $\sigma\in\{15,25,50, 75, 100\}$ (Table~\ref{table:sync_bsd500_results}).
Images in BSD500 are high-resolution and variable in size; for training we
sample random $128\times128$ RGB patches (after resizing if necessary to ensure
a minimum side length of 128). Noise is injected \emph{on-the-fly} during
training with the target $\sigma$, and $\sigma$-aware models receive a
constant single-channel $\sigma$-map ($\sigma/255$) concatenated to the input.
At test time, full images are processed by dividing them into overlapping
$128\times128$ crops with reflection padding and center blending to avoid seam
artifacts; denoised crops are then stitched back to form the final output.
Performance is reported as mean PSNR and SSIM over the 200 test images. 

\begin{figure}[ht]
\centering
\begin{tabular}{cccc}
\includegraphics[width=0.23\linewidth]{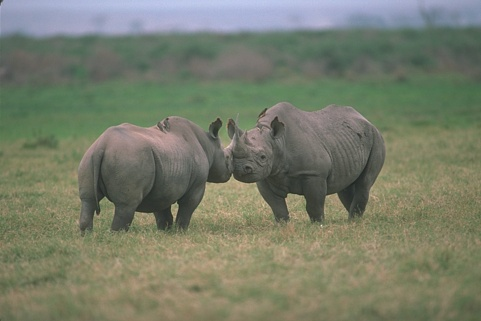} &
\includegraphics[width=0.23\linewidth]{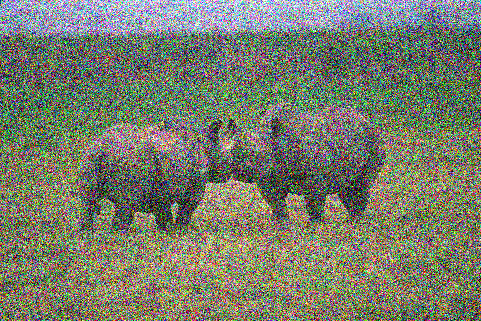} &
\includegraphics[width=0.23\linewidth]{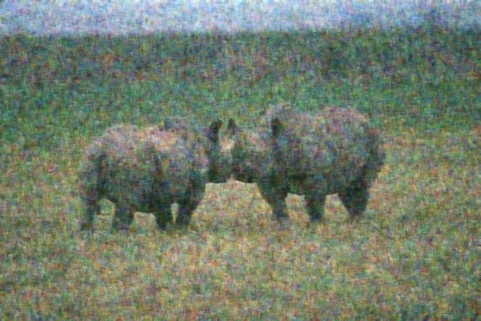} &
\includegraphics[width=0.23\linewidth]{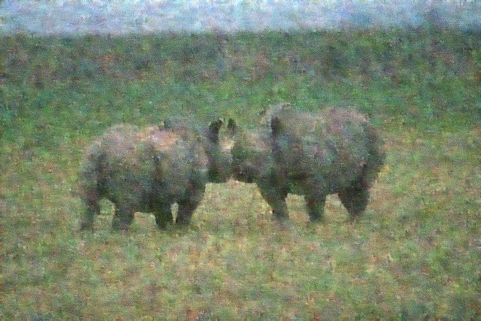} \\
(a) Clean & (b) Noisy ($\sigma$=75) & (c) UNet Denoised & (d) WIPUNet Denoised \\[0.5em]

\includegraphics[width=0.23\linewidth]{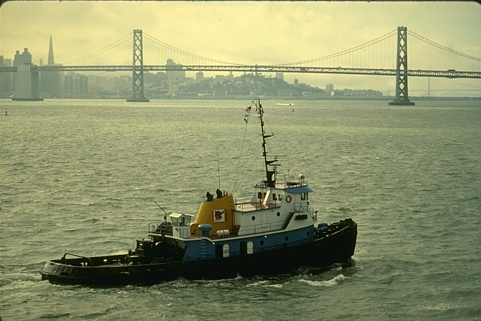} &
\includegraphics[width=0.23\linewidth]{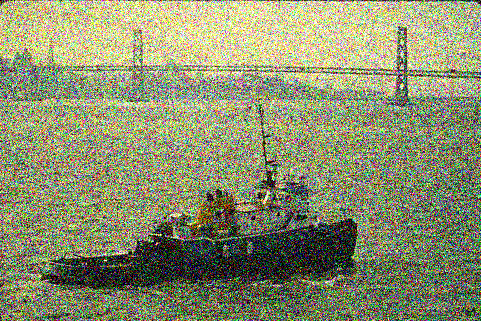} &
\includegraphics[width=0.23\linewidth]{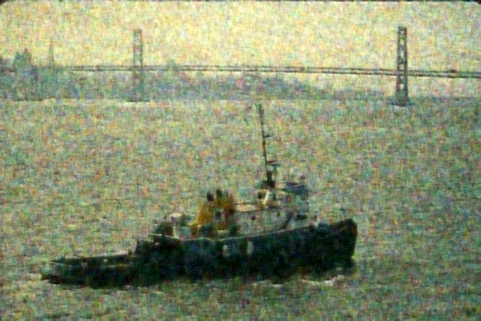} &
\includegraphics[width=0.23\linewidth]{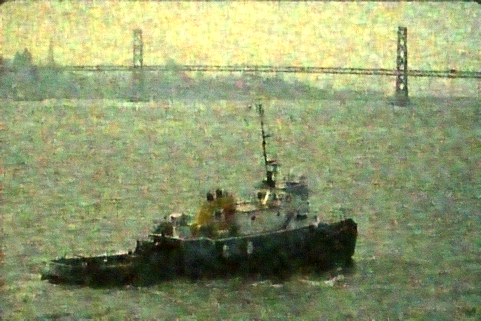} \\
(e) Clean & (f) Noisy ($\sigma$=75) & (g) UNet Denoised & (h) WIPUNet Denoised \\[0.5em]

\includegraphics[width=0.23\linewidth]{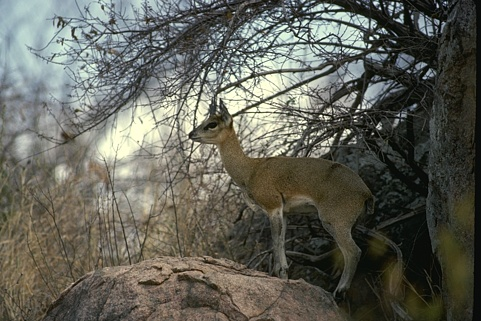} &
\includegraphics[width=0.23\linewidth]{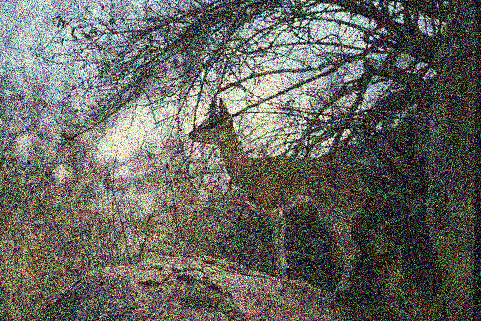} &
\includegraphics[width=0.23\linewidth]{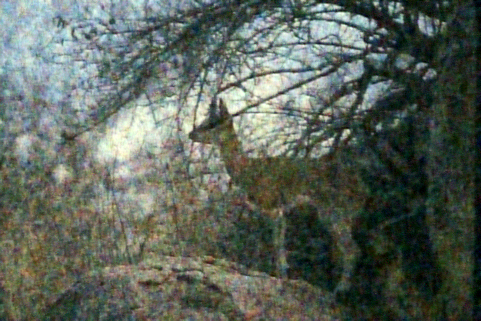} &
\includegraphics[width=0.23\linewidth]{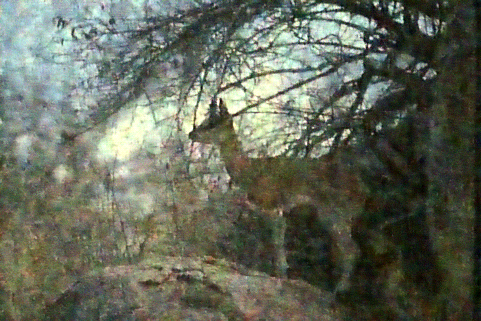} \\
(i) Clean & (j) Noisy ($\sigma$=75) & (k) UNet Denoised & (l) WIPUNet Denoised \\
\end{tabular}
\caption{Visual comparison of denoising results at noise level $\sigma$=75. The first column shows clean reference images, the second column shows noisy inputs, and the third and fourth columns show denoising results from UNet and WIPUNet models, respectively. Rows correspond to different test images from the BSD500 dataset.}
\label{fig:denoising_comparison}
\end{figure}

WIPUNet remains competitive with UNet at $\sigma=15$ and $\sigma=25$, and surpasses UNet at $\sigma=50$ (25.19 vs.\ 24.85 dB PSNR). At higher noise levels, $\sigma=75$ and $\sigma=100$, WIPUNet continues to outperform UNet (23.57 vs.\ 23.00 dB and 23.10 vs.\ 21.86 dB, respectively), while also maintaining a clear advantage over RestormerLite, DnCNN, and FFDNet. Notably, the performance gap between WIPUNet and UNet widens as noise increases — from just 0.3 dB at $\sigma=50$ to over 1.2 dB at $\sigma=100$. This trend highlights the robustness of physics-inspired priors under severe corruption. These results corroborate the trend observed on CIFAR-10: physics-inspired priors become increasingly beneficial as corruption strengthens, improving stability when purely data-driven baselines degrade. Figure~\ref{fig:denoising_comparison} illustrates this effect, showing denoising comparisons on BSD images corrupted with Gaussian noise at $\sigma=75$.

\paragraph{Overall implications.}
Taken together, the results across both CIFAR-10 and BSD500 demonstrate that WIPUNet delivers not only strong performance but also \emph{scaling robustness}: the more challenging the noise regime, the more useful its physics-inspired modules become. On CIFAR-10, WIPUNet matched or slightly exceeded UNet under mild and moderate noise, and the advantage steadily widened as corruption intensified — surpassing UNet by 0.5 dB at $\sigma=50$, 0.6 dB at $\sigma=75$, and 0.7 dB at $\sigma=100$. On BSD500, the same trend was observed: WIPUNet surpassed UNet beginning at $\sigma=50$ and maintained a clear lead at $\sigma=75$ and $\sigma=100$, while consistently outperforming weaker baselines such as DnCNN and RestormerLite across all regimes. These cross-dataset findings reinforce the central message: the value of WIPUNet lies less in setting new benchmarks than in demonstrating feasibility. Physics-inspired inductive biases can match or exceed standard CNN baselines even in non-physics domains, while offering interpretability and resilience that purely empirical designs lack.

\section{Related Work}

\paragraph{Image denoising.} 
Classical CNN-based denoisers such as DnCNN~\cite{zhang2017beyond} and FFDNet~\cite{zhang2018ffdnet} 
established residual learning and noise-level conditioning for AWGN. 
UNet~\cite{ronneberger2015unet} and variants are strong low-level backbones, 
while Restormer~\cite{zamir2022restormer} scales attention to high-resolution restoration. 
More recent work explores frequency-space augmentation and adaptive noise modeling 
(e.g., adversarial frequency mixup~\cite{ryou2024afm}, LAN~\cite{kim2024lan}), and ultra-efficient 
lookup-table approaches (DnLUT~\cite{yang2025dnlut}). 
Our study is orthogonal: we do not target SOTA; instead, we examine whether \emph{physics-inspired priors} 
yield robustness gains that complement these trends.

\paragraph{Physics-inspired priors.} 
In high-energy physics, pileup mitigation strategies 
\cite{ATLAS2015_PileupMitigation,Berta2019_ICS,CMS2020_PileupMitigation,PUPPI2014,Cerri2018_GNN} 
encode conservation, conditioning, and isolation. Recent ML-based approaches in HEP have begun embedding such priors for tasks including vertex finding, calorimeter denoising, and jet reconstruction. 
Our work bridges these directions by importing pileup-inspired strategies into computer vision, yielding interpretable architectures that remain 
competitive with strong CNN baselines and become increasingly beneficial as corruption intensifies.

\section{Limitations and Future Work}

While our results demonstrate that physics-inspired priors provide competitive and robust denoisers, several limitations remain. First, although we evaluated both on CIFAR-10 and BSD500, these datasets are relatively small; scaling to larger and more diverse benchmarks such as ImageNet or scientific imaging data will be necessary to fully establish generalization. Second, we restricted attention to additive white Gaussian noise; more realistic corruptions (structured, spatially varying, or signal-dependent noise) remain unexplored but are especially relevant to physics, astronomy, and medical applications. Third, our evaluation metrics were limited to PSNR and SSIM; incorporating perceptual measures or human preference studies could provide complementary insights into reconstruction quality. 
Finally, while WIPUNet consistently approaches or surpasses strong baselines under high noise, it does not universally exceed vanilla UNet, suggesting that additional architectural refinements or hybridization with transformer-style models may be needed. We also did not re-train the latest SOTA systems (e.g.,~\cite{ryou2024afm,kim2024lan,yang2025dnlut}); this is intentional given our proof-of-concept focus. We deliberately kept the scope narrow to emphasize the physics–vision connection, rather than exhaustively benchmarking architectures.

Future work can explore extensions in three directions: (i) applying physics-inspired priors to structured and non-Gaussian noise, (ii) integrating physics-inspired inductive biases into transformer architectures to couple domain priors with long-range modeling capacity, and (iii) evaluating cross-dataset transfer to probe domain shift. Public release of pretrained models may further enable the community to explore physics-guided architectures for a wide range of restoration tasks.

Finally, many physics studies now represent jets and events as images; a practical extension is to apply WIPUNet to such physics image modalities (jet images, calorimeter maps, track $\eta$–$\phi$ occupancy) to test robustness under realistic detector conditions.

\section{Conclusion}
We translated pileup-mitigation principles from collider physics into modular inductive biases for image denoising 
and integrated them into \textbf{WIPUNet}. Evaluations across five Gaussian 
noise levels ($\sigma=15,25,50, 75, 100$) on CIFAR-10, and complementary experiments on BSD500, 
demonstrate that WIPUNet consistently outperforms classical CNN denoisers and even 
surpasses the strong UNet baseline as noise level intensifies. Notably, WIPUNet maintains parity with UNet at low noise, modestly exceeds it at moderate noise, and achieves a steadily widening margin under severe noise while outperforming UNet. On BSD500, a similar trend is observed. These findings reinforce the conclusion that physics-inspired priors provide robustness precisely where conventional CNNs deteriorate. The central message is that the advantage of WIPUNet is not in peak performance at low noise, but in the \emph{widening robustness gap under severe corruption}, where physics-inspired priors increasingly stabilize learning as conventional baselines collapse.

This work is a \textbf{proof of concept}: the objective is not SOTA scores, but demonstrating that conservation, 
noise conditioning, isolation, and multi-scale fusion offer interpretable, domain-inspired stability under heavy noise. 
We expect these priors to be complementary to modern denoisers and particularly valuable for structured, non-stationary 
noise and can be useful for applications in high-energy physics, 
astronomy, and medical imaging. Future work can target non-Gaussian and spatially varying noise and 
combine physics priors with transformer-style long-range modeling.

\section*{Acknowledgments}

 The author is thankful to Prof. Ariel Schwartzman of SLAC National Accelerator Laboratory-Stanford University for extensive discussions on Pileups in particle physics in general. Author is supported by DE-SC0017647. 

\bibliographystyle{unsrt}

\end{document}